\documentclass[letterpaper]{article} 
\usepackage{aaai23}  
\usepackage{times}  
\usepackage{helvet}  
\usepackage{courier}  
\usepackage[hyphens]{url}  
\usepackage{graphicx} 
\usepackage{natbib}  
\usepackage{caption} 
\usepackage{algorithm}
\usepackage{algorithmic}

\usepackage{amsmath}
\usepackage{xcolor}
\usepackage{amsfonts}
\usepackage{array}
\usepackage{multirow}
\usepackage{booktabs}
\usepackage{url}
\usepackage{appendix}

\usepackage{newfloat}
\usepackage{listings}
\DeclareCaptionStyle{ruled}{labelfont=normalfont,labelsep=colon,strut=off} 
\floatstyle{ruled}
\newfloat{listing}{tb}{lst}{}
\floatname{listing}{Listing}
\title{Reducing Domain Gap in Frequency and Spatial domain for Cross-modality Domain Adaptation on Medical Image Segmentation}

\author{
    Shaolei Liu\equalcontrib,
    Siqi Yin\equalcontrib,
    Linhao Qu,
    Manning Wang\thanks{Corresponding Authors}
}
\affiliations{
    Digital Medical Research Center, School of Basic Medical Science, Fudan University, Shanghai 200032, China\\
    Shanghai Key Lab of Medical Image Computing and Computer Assisted Intervention\\
    \{slliu, lhqu20, mnwang\}@fudan.edu.cn, sqyin21@m.fudan.edu.cn
}

\usepackage{bibentry}

\begin{document}

\maketitle

\begin{abstract}
    Unsupervised domain adaptation (UDA) aims to learn a model trained on source domain and performs well on unlabeled target domain. 
    In medical image segmentation field, most existing UDA methods depend on adversarial learning to address the domain gap between different 
    image modalities, which is ineffective due to its complicated training process. In this paper, we propose a simple yet effective UDA method 
    based on frequency and spatial domain transfer under multi-teacher distillation framework. In the frequency domain, we first introduce 
    non-subsampled contourlet transform for identifying domain-invariant and domain-variant frequency components (DIFs and DVFs), and then 
    keep the DIFs unchanged while replacing the DVFs of the source domain images with that of the target domain images to narrow the domain 
    gap. In the spatial domain, we propose a batch momentum update-based histogram matching strategy to reduce the domain-variant image style bias. 
    Experiments on two cross-modality medical image segmentation datasets (cardiac, abdominal) show that our proposed method achieves superior 
    performance compared to state-of-the-art methods.
\end{abstract}

\section{Introduction}
Deep learning has a wide range of applications in the medical field, including image segmentation \cite{1,2}, classification \cite{3,4}, 
detection \cite{5,6}, etc. The success of deep learning heavily depends on large amount of annotated data for model training, but the pixel-level 
annotation of medical images for segmentation is expensive and time-consuming, because the annotation usually can only be done by experienced radiologists.
 To tackle the difficulty of obtaining new annotated datasets, one solution is to utilize other annotated datasets \cite{8,9}. However, the 
 differences on imaging principles and equipment parameters (scanners, protocols and modalities) lead to 
 domain gap between datasets, and the performance of a model trained on one domain tends to deteriorate when it is directly used in another 
 domain \cite{2,10,11}. Unsupervised domain adaptation (UDA) is the technique to address this problem by training a network with labeled 
 source domain data and unlabeled target domain data, aiming at improving the performance on target domain \cite{12}.

The main objective of UDA is to minimize the influence of domain gap on the performance of the trained model in the target domain, 
which can be realized by either generating target-like images for model training or encouraging the model to focus on domain-invariant 
information instead of domain-variant information. In the UDA for medical image segmentation, most studies leverage adversarial learning 
to stylize source-domain images at the image level \cite{1,2,13,14,16,17,18,41} to generate target-like images for model training. Other studies 
apply adversarial learning at the feature level to maximize the confusion between representations in source and target domains to 
make the model learn domain invariant representations \cite{20,21}. Although satisfactory results have been achieved, the adversarial training 
process  is complicated and is prone to collapse.

In recent years, several UDA methods without using adversarial learning have been developed in the CV field, which can be 
divided into two categories: spatial domain-based methods and frequency domain-based methods. Spatial domain-based methods generate target-like images 
by simple cutmix-like region replacement \cite{22}, statistical information adjustment \cite{23,24} 
or histogram matching \cite{26}. Frequency domain-based methods first 
transform images into frequency components by Discrete Fourier Transform (DFT) \cite{19,26} or Discrete Cosine Transform (DCT) \cite{27} and narrow 
domain gap by manipulating the frequency components. These methods have yielded promising results in the CV field, 
but no studies have been explored in medical image analysis. 

In this paper, we propose a novel UDA method in the frequency domain and combine it with a spatial domain UDA method under a multi-teacher 
distillation framework to achieve better segmentation performance in the target domain. Specifically, we introduce Non-Subsampled Contourlet 
Transform (NSCT) for the first time in frequency domain-based UDA. Compared to DFT and DCT, NSCT can produce finer, anisotropic, and multi-directional 
frequency components and reduce spectrum overlapping. We identify the domain-variant frequency components (DVFs) and domain-invariant frequency components (DIFs) 
in all NSCT frequency components and replace the DVFs of source 
domain images with that of target domain images while keeping the DIFs unchanged. Thus, the negative effects of the DVFs of source domain images are 
reduced in model training. In the spatial domain, we propose a batch momentum update-based histogram matching strategy to align the image styles. 
Finally, we train two segmentation models by using the above frequency domain UDA and spatial domain UDA strategies and use them as teachers to train 
a student model for inference in a multi-teacher distillation framework. Our method 
outperforms existing UDA methods in experiments on cross-modality cardiac and abdominal multi-organ datasets. The main contributions are summarized as follows:

\begin{itemize}
    \item We introduce NSCT to perform frequency domain-based unsupervised domain adaptation for the first time.
    \item We propose a simple and effective UDA framework based on multi-teacher distillation to integrate 
    frequency-domain and spatial-domain UDA strategies to further enhance the model's UDA performance.
    \item Extensive experiments show that our method outperforms the state-of-the-art methods on cardiac and abdominal multi-organ segmentation datasets.
\end{itemize}

\section{Related Work}
Existing studies can be mainly divided into adversarial learning-based and non-adversarial learning-based methods.

\subsection{Adversarial Learning-based Methods}
Most of the UDA studies for medical image segmentation use adversarial learning to align domain-variant information to reduce domain gap \cite{1,2,13,14,16,17,18,41}, 
which can be further divided into image-level methods and feature-level methods. In image-level methods, Generative Adversarial Networks (GANs) or its variant 
CycleGAN \cite{17} are used for generating images with target-like appearance from source domain image for model training \cite{1,2,13,14,16,17,18,41}. Adversarial learning 
can also be used in the feature level to extract domain-invariant features to improve the model generalization on the target domain \cite{2,13}. The main problem of this kind 
of methods lies in the complicated training process and the difficulty in network convergence.

\subsection{Non-adversarial Learning-based Methods}
A series of methods without adversarial learning have been proposed to achieve UDA in a simple way in the CV field. 
These methods can be further classified into spatial domain-based and frequency domain-based methods.

\noindent \textbf{2.2.1 Spatial Domain-based Methods} 

\noindent
Spatial domain-based methods directly adjust the source domain images or their features to narrow the gap 
with the target domain. For example, Nam et al. \cite{24} used adaptive instance normalization to replace the style information revealed in channel-wise mean 
and standard deviation of the source domain images with that of randomly selected target domain images for style transfer before training. Some 
studies used histogram matching \cite{25} or mean and variance adjustment \cite{23} in the LAB color space to reduce domain gap. 

\noindent \textbf{2.2.2 Frequency Domain-based Methods} 

\noindent
Frequency domain-based methods first decompose the source and target domain images into frequency components 
(FCs) and narrow the domain gap by manipulating the FCs. For example, in the studies of Yang et al. \cite{19} and Xu et al. \cite{26}, DFT is used for decomposition and the low frequency part of the 
amplitude spectrum of source images is replaced with that of target domain images to generate target-like images. In Huang et al. \cite{27}, DCT is 
used to decompose images into FCs, which are then divided into two categories: domain-variant FCs and domain-invariant FCs. Then the domain-variant 
FCs of a source image is replaced with the corresponding FCs of a randomly-selected target domain image, and a target-like image is reconstructed by 
inverse transformation for model training. However, the FCs of DCT is in a single decomposition scale and the spectral energy is 
mainly concentrated in the upper left corner including a little low-frequency information. In this paper, we adopted NSCT for decomposition, which 
can obtain multi-scale, multi-directional and anisotropic FCs.

\section{Method}

\subsection{Problem Formulation}

\begin{figure*}[htbp]
    \centering
    \includegraphics[scale=0.53]{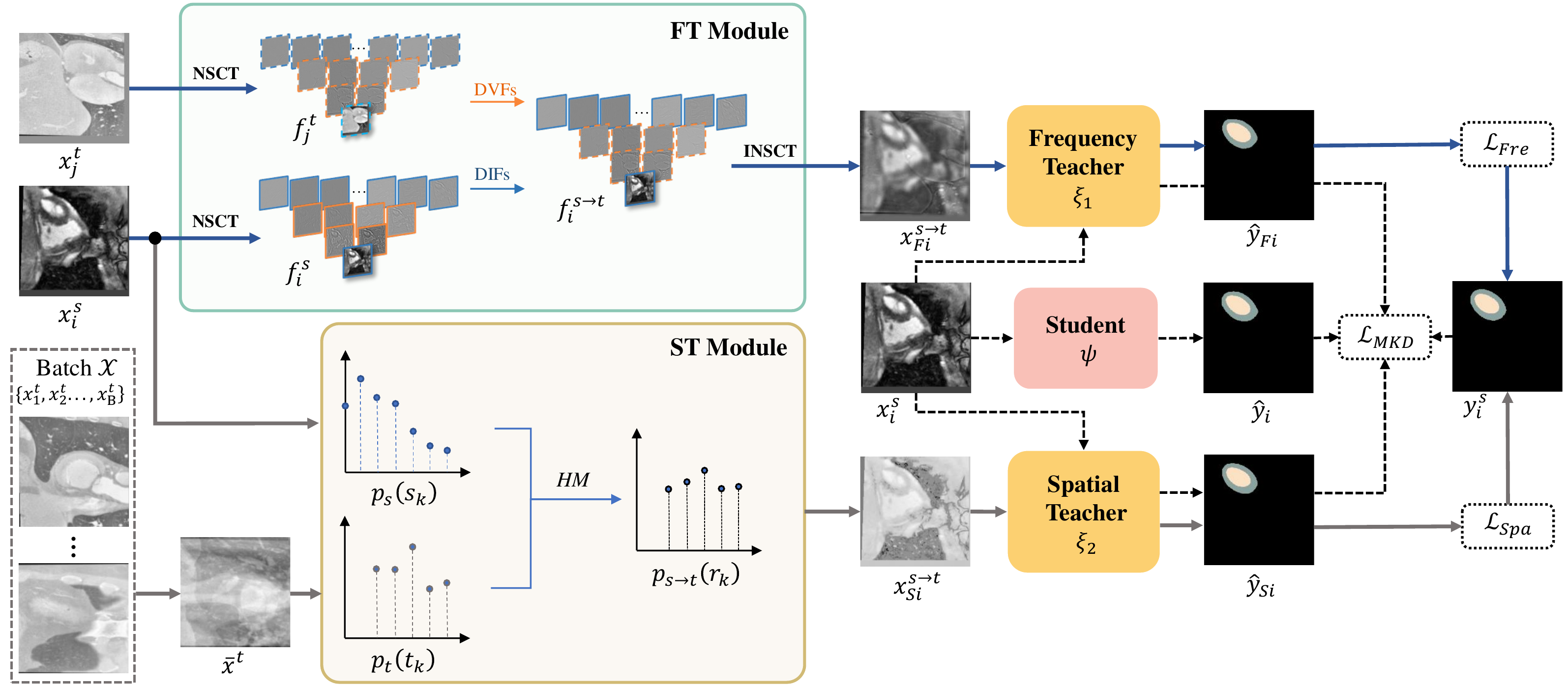}
    \caption{The training framework of our proposed method. The solid line represents the training process of the teacher networks, and the dashed line 
    represents the training process of the multi-teacher distillation network.}
    \label{figure1}
\end{figure*}

Given source domain data $D_s={(x_i^s,y_i^s)}_{i=1}^{N_s}$, where $x_i^s\in{\mathbb{R} ^{H\times{W}}}$ represents a source medical image with segmentation label 
 $y_i^s\in{\mathbb{R}^{{n_c}\times{H}\times{W}}}$, and $n_c$ denotes the number of segmentation categories. Similarly, $D_t={(x_j^t)}_{j=1}^{N_t}$ is the target domain data 
without segmentation labels. The objective of this paper is to train a segmentation model with $D_s$ and $D_t$ that performs well on the target domain.

The overall framework proposed in this paper is shown in Figure \ref{figure1}, where $x_i^s$ and $x_j^t$ are a source domain image and a target domain image, 
respectively, and $\bar{x}^t$ is the average image of a batch of target domain images. We use a frequency domain transfer (FT) module and a spatial 
domain transfer (ST) module to generate two sets of target-like images $x_{Fi}^{s\to{t}}$ and $x_{Si}^{s\to{t}}$, respectively. Each image set is used 
to train a teacher model, and the two teachers are then utilized to train a student model in a 
multi-teacher distillation framework for inference. The FT module takes $x_i^s$ and $x_j^t$ as inputs and each of them are decomposed into 15 frequency components 
(FCs) using NSCT, which are denoted as $f_i^s$ and $f_j^t$, respectively. We empirically identify domain invariant frequency components (DIFs) and 
domain variant frequency components (DVFs) in the NSCT components. The DVFs in $f_i^s$ are replaced with that in $f_j^t$, and the resulted source domain FCs are used 
to generate $x_{Fi}^{s\to{t}}$ using inverse NSCT. Details of the FT module are presented in Section 3.2.

In the ST module, we first calculate the normalized histogram $p_s(s_k)$ of the source domain image $x_i^s$. At the same time, we 
randomly sample a batch of target domain images $\mathcal{X}  =\{x_1^t,x_2^t…,x_B^t\}\in{D_t}$ and calculate their average image $\bar{x}^t$ and its normalized histogram $p_t(t_k)$. 
Then, $p_s(s_k)$ is aligned with $p_t(t_k)$ by histogram matching to obtain the transferred source domain image $x_{Si}^{s\to{t}}$. In this process, 
we introduce momentum updates to $p_t(t_k)$ to make it change steadily. Details of the ST module are shown in Section 3.3. 

We use $x_{Fi}^{s\to{t}}$ and $x_{Si}^{s\to{t}}$ to train two teacher models 
$\xi_1$ and $\xi_2$, respectively. Then, we utilize  an entropy-based multi-teacher distillation strategy \cite{34} to integrate the information 
learned by the two teachers to train a student model for inference (see the dashed arrow data flow in Figure \ref{figure1}). Details of the distillation framework are 
presented in Section 3.4.

\subsection{Frequency Domain Transfer Module}
In this section, we will first briefly introduce NSCT for completeness and then present how it is used for frequency domain image transfer.

\noindent \textbf{3.2.1 Non-Subsampled Contourlet Transform (NSCT)}

\noindent NSCT can efficiently capture multi-scale visual information of images by decomposing an image into 
high-pass sub-bands representing details and low-pass sub-bands representing structural information \cite{29,33}. Concretely, NSCT consists 
of two parts: a Non-Subsampled Pyramid (NSP) and a Non-Subsampled Directional Filter Bank (NSDFB), which are used to obtain multi-scale 
and multi-directional frequency components, respectively. We briefly introduce the three-layer NSCT used in this study here and more 
details can be found in the \textbf{Supplementary Material Section 2}.

\begin{figure}[htbp]
    \centering
    \includegraphics[scale=0.28]{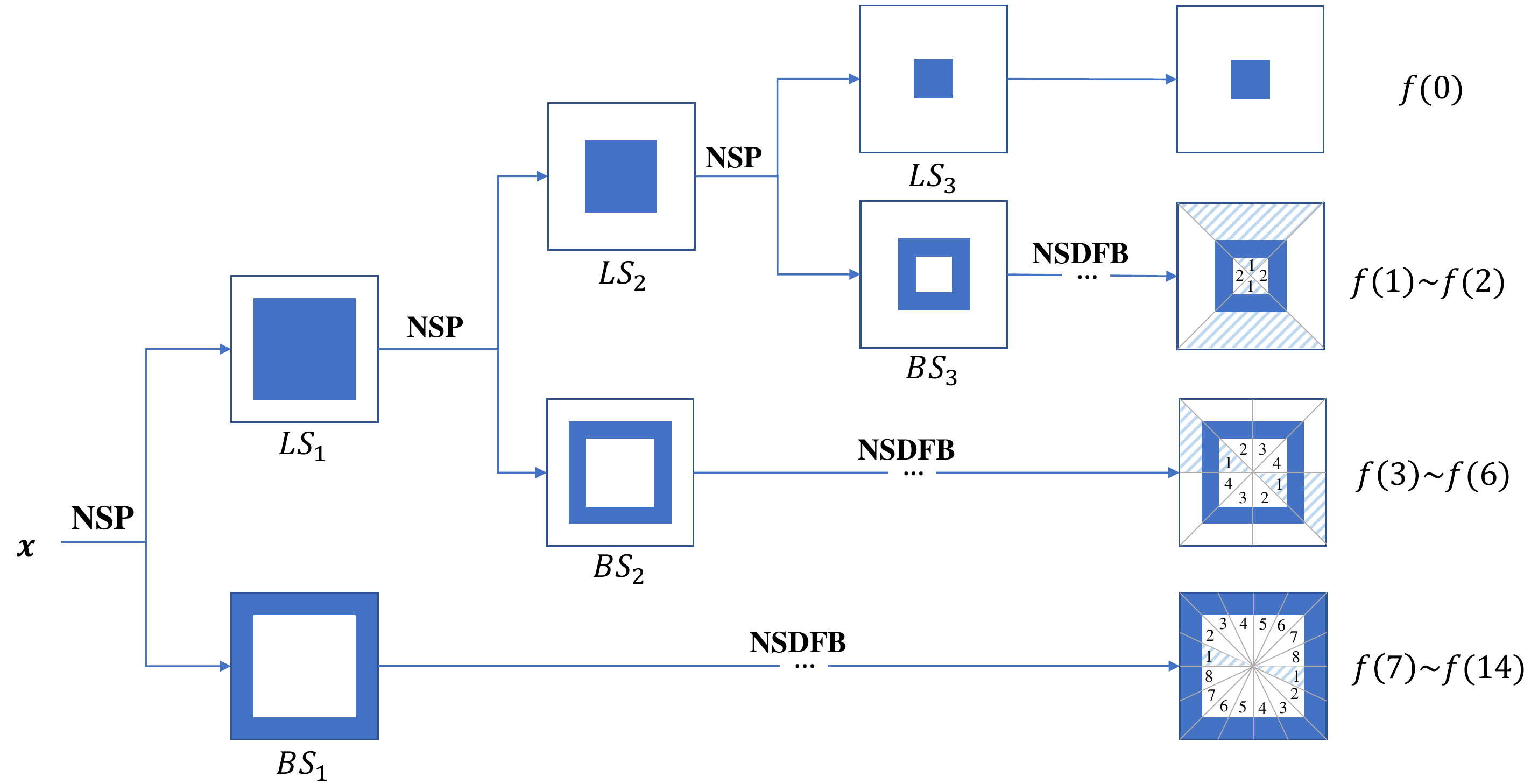}
    \caption{The detail of three-layer NSCT decomposition.}
    \label{figure2}
\end{figure}

\begin{figure}[!b]
    \centering
    \includegraphics[scale=0.45]{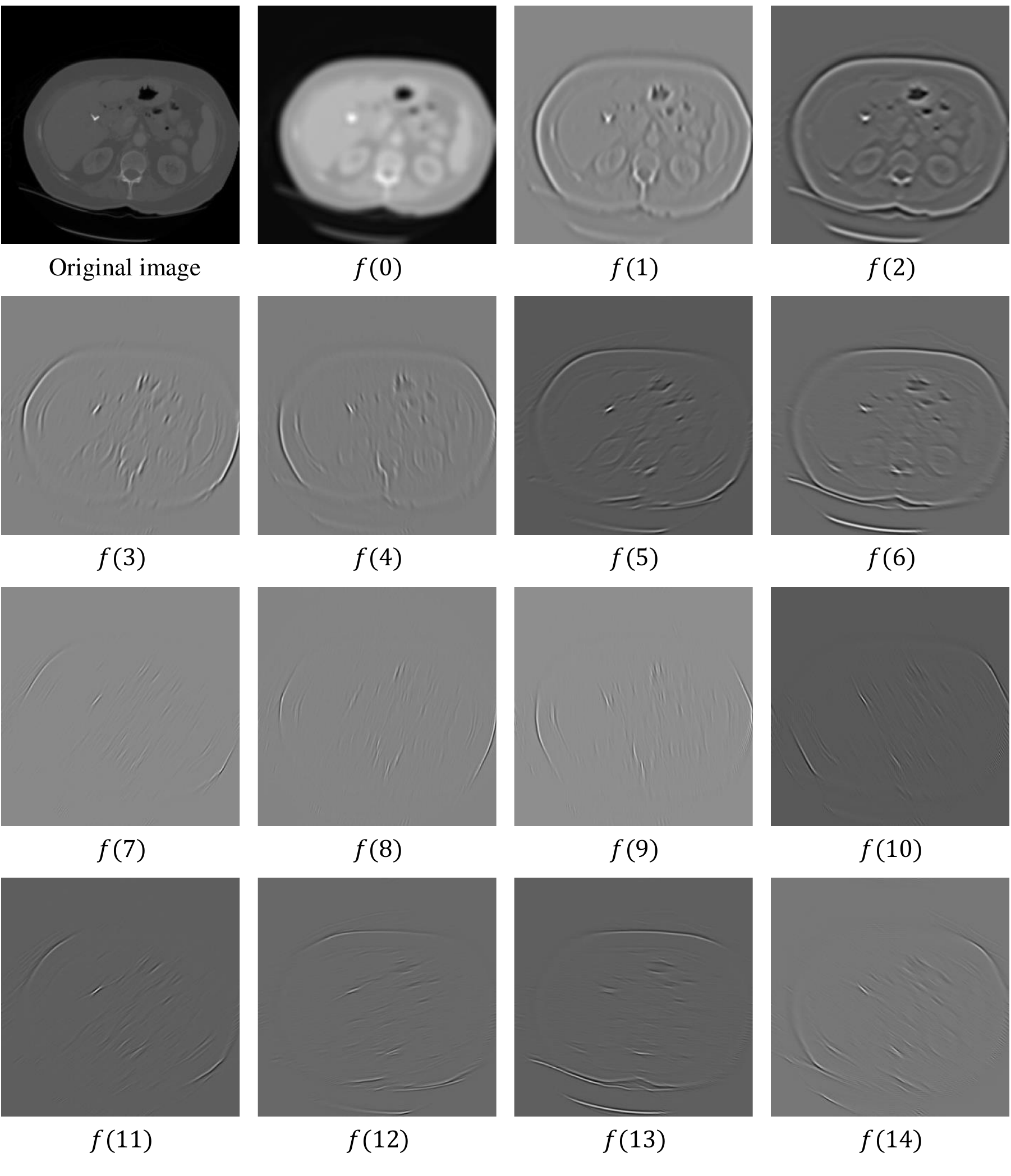}
    \caption{Example of NSCT decomposition of a abdominal MRI image.}
    \label{figure3}
\end{figure}
As shown in Figure \ref{figure2}, for an input image $x$, a low-frequency sub-band image $LS_1$ and a high-frequency sub-band image $BS_1$ 
are obtained after NSP decomposition. The low-frequency sub-bands at the $l^{th}$ layer $LS_{l}$ will continue to be decomposed by 
NSP to obtain $LS_{l+1}$ and $BS_{l+1}$. Then, the high-frequency sub-bands $BS_l$ are further decomposed into $2^l$ multi-directional 
frequency sub-bands using NSDFB. After the $L$-level decomposition, the input image $x$ is transformed into a series of FCs, 
including the low-frequency sub-band $f(0)$, and a total number of $\sum_{l=1}^{L}2^l$ high-frequency sub-bands 
$f(1)\sim f(\sum_{l=1}^{L}2^l)$.

Taking an abdominal MRI image $x_i^s$ as an example, we obtain its FCs $f_i^s$ by three-layer NSCT decomposition:

{\small
\begin{equation}
    f_i^s={NSCT}(x_i^s)=\{f_i^s (0),f_i^s (1),…,f_i^s (14)\} \label{eq1}
\end{equation}
}

The visualization of each sub-bund FC is shown in Figure \ref{figure3}. We can see that the FC of low frequency sub-band retain 
most of the semantic content of the image, while the FCs corresponding to high frequency sub-bands represent the 
structure and texture of the image in different directions.

\noindent \textbf{3.2.2 NSCT-based Frequency Transfer Strategy}

\noindent In order to narrow the domain gap using frequency components, we propose an NSCT-based frequency domain transfer 
strategy between the two domains.

First, for each source domain image, we decompose it into 15 FCs using NSCT, and then reconstruct a series of images 
from different combinations of these FCs. We train a baseline model using the original source domain images and train 
a transfer model using the images reconstructed from each FCs combination. Then DVFs and DIFs are identified according 
to the performance of all these models on a synthetic target domain data. Specifically, an improved performance over 
the baseline model implies that the rejected FCs in the corresponding combination are domain-variant, and removing 
these FCs can prevent the model from learning too many domain-variant features to facilitate domain adaptation. 
On the contrary, a decreased performance over the baseline model implies that the removed FCs 
are domain-invariant, and by retaining these components, the model is encouraged to learn domain-invariant features to 
facilitate domain adaptation.

Due to page limitation, the experimental details of identifying DIFs and DVFs are given in \textbf{Supplementary Material Section 3}, 
and we only give the results here, i.e. the FCs combination with the low-frequency $f_i^s (0)$ and the third-layer high-frequency 
components $f_i^s (7\sim 14)$ achieve the best results. Therefore, we consider the low-frequency FC and the third-layer FCs as DIFs, and 
the other FCs as DVFs.

After identifying the DVFs and DIFs, for a source domain image $x_i^s$, we keep its DIFs unchanged and replace its 
DVFs with that of a randomly selected target domain image to obtain the adapted FCs $f_{i}^{s\to{t}}$. Then inverse NSCT 
is applied on $f_{i}^{s\to{t}}$ to obtain the corresponding adapted image $x_{Fi}^{s\to{t}}$, which is used to train the frequency 
domain transfer-based segmentation teacher network $\xi_1$ with the loss function $ \mathcal{L}_{Fre}$  defined as follows:

{\small
\begin{align}
    \mathcal{L}_{Fre}=\sum \left[ \mathcal{L}_{CE} \left(\hat{y}_{Fi},y_i^s\right)+ \mathcal{L}_{Dice} \left(\hat{y}_{Fi},y_i^s\right) \right] 
    \label{eq2}
\end{align}
}

\noindent
where $\mathcal{L}_{CE}$ is the cross-entropy loss and $\mathcal{L}_{Dice}$ is the Dice loss. $\hat{y}_{Fi}$ is the predicted segmentation result.

\subsection{Spatial Domain Transfer Module}
In order to further reduce the domain gap, we utilize the spatial domain transfer module to generate another 
target-like image for model training by histogram matching \cite{35}. We propose to use batch momentum update to 
calculate an average histogram of a batch of target domain images for matching, rather than the entire target 
domain, which tends to smooth the target domain histogram and ignores the intensity variants \cite{36}.

Concretely, given the source domain image $x_i^s$ and a batch of target domain image $\mathcal{X}=\{x_1^t,x_2^t…,x_B^t\}$, 
we first normalize all these images to make them have integer intensity values falling in the range of $[0,L-1]$. 
The normalized histogram of $x_i^s$ is defined as $p_s(s_k)=n_k/{HW}$, where  $s_k,k=0,1,…L-1$ represents the $k^{th}$ 
intensity value, and $n_k$ is the number of pixels with intensity $s_k$. Similarly, the histogram of the average target 
image of the current batch $\bar{x}^t  =1/B \sum_{i=1}^{B}x_i^t$  is denoted as $p_t(t_k)$. 

Then, we use the source image histogram to find the histogram equalization transformation $T$ as follows, and 
round the obtained value ${\mu}_k$ to the integer range $[0,L-1]$.

{\small  
\begin{align}
    {\mu}_k=T\left(s_k\right)=\left(L-1\right) \sum_{j=0}^{k}{p_s(s_j)}=\frac{L-1}{HW}\sum_{j=0}^{k}{n_j}\label{eq3}
\end{align}
}

\noindent
where $k=0,1,…,L-1$, $n_j$ is the number of pixels with intensity value $s_j$. 
Similarly, we can use the histogram of the average target intensity image $p_t(t_k)$ to calculate its histogram 
equalization transformation function $G(t_q),q=0,1,…L-1$. Then round the resulting values $G(t_q)$ to the integer range 
$[0,L-1]$ and store the values in a table.

{\small  
\begin{equation}
    G(t_q)=(L-1) \sum_{i=0}^{q}p_t(t_i)\label{eq4}
\end{equation}
}

For every value of $\mu_k$, we use the stored values of $G$ to find the corresponding value of $t_q$ so that $G(t_q)$ 
is closest to $\mu_k$ and then we store these mappings from $\mu$ to $t$. Note that, if there are more than one values 
of $t_q$ satisfies the given $\mu_k$, we will choose the smallest value. Finally, we use the found mapping to map 
each equalized source domain image pixel value $\mu_k$ to the corresponding pixel with value $t_q$ 
and we can obtain the final histogram-matched image $x_{Si}^{s\to{t}}$.

In addition, for the update of the average intensity image of the current batch in the target domain, we use 
the cumulative average based on batch momentum update to slowly change the current average image. 

{\small  
\begin{align}
    \bar{x}_i^t=\eta\bar{x}_i^t+(1-\eta)\bar{x}_{i-1}^t\label{eq5}
\end{align}
}

\noindent
where $\eta$ is set as 0.7, $\bar{x}_i^t$ is the average target image of current batch $i$ and $\bar{x}_{i-1}^t$ is that of last batch. Finally, we input $x_{Si}^{s\to{t}}$ for training the second segmentation teacher 
network $\xi_2$, and the loss function $\mathcal{L}_{Spa}$ is defined as follows.

{\small  
\begin{align}
    \mathcal{L}_{Spa}=\sum\left[ \mathcal{L}_{CE}\left(\hat{y}_{Si},y_i^s\right)+ \mathcal{L}_{Dice} \left(\hat{y}_{Si},y_i^s \right)\right]\label{eq6}
\end{align}
}

\noindent
where $\hat{y}_{Si}$ represents the predicted segmentation results.

\subsection{Multi-teacher Knowledge Distillation for Domain Adaptation}

Given the source domain data $D_s={(x_i^s,y_i^s)}_{i=1}^{N_s}$ and two trained teacher models  $\xi_1$ and $\xi_2$, 
we train a student model for inference using the entropy-based 
dynamic multi-teacher distillation strategy \cite{34}. As shown in Figure \ref{figure1}, we first 
input a source domain image $x_i^s$ into $\xi_1$ and $\xi_2$ to obtain the predicted 
segmentation category probability maps $\hat{y}_{Fi}$ and $\hat{y}_{Si}$, respectively. Then, we compute the entropy-based distillation 
loss to train the student network $\psi$ with $x_i^s$ and predict its probability maps $\hat{y}_{i}$. 
The loss function for training the student network is:

{\small  
\begin{align}
    &\mathcal{L}_{MKD}=\sum_{i=1}^{N_s}\{\alpha\left(\mathcal{L}_{CE}\left(\hat{y}_{i},y_i^s\right)+\mathcal{L}_{Dice}\left(\hat{y}_{i},y_i^s \right)\right)  \notag\\
   & + \omega_{KD}^{F}\mathcal{L}_{KD} \left(\hat{y}_{Fi},y_i^s \right) + \omega_{KD}^{S}\mathcal{L}_{KD} \left(\hat{y}_{Si},y_i^s \right)\}   \label{eq7} 
    \end{align}
}
{\small
\begin{align}
    \mathcal{L}_{KD} &=\frac{1}{HW}\sum_{h=1}^{H}\sum_{w=1}^{W}KL\left(\sigma\left(\delta/\tau\right)||\sigma\left(\hat{y}_{i}/\tau\right)\right), \notag \\
     \omega_{KD}^{F}& = 1 - \frac{H(\hat{y}_{Fi})}{H(\hat{y}_{Fi})+H(\hat{y}_{Si})}, \notag \\
     \omega_{KD}^{S}& = 1 - \frac{H(\hat{y}_{Si})}{H(\hat{y}_{Fi})+H(\hat{y}_{Si})}.
    \label{eq8}
\end{align}
}



\noindent
where $\delta=\hat{y}_{Fi},\hat{y}_{Si} $, $\sigma$ represents the softmax function, $KL$ denotes the Kullback-Leibler divergence, and $\tau$ is a temperature. $\alpha$ 
is a trade-off hyperparameter, which is set as 10. $H(\cdot)$ represents the entropy computation.

\section{Experiments Results}

\subsection{Implementation Details}

\noindent  \textbf{4.1.1 Datasets and Implementation Details}

\noindent \textbf{Abdominal Multi-organ Segmentation.} We use the same Abdominal Multi-Organ dataset as SIFA \cite{2}, in which 
the publicly available CT data \cite{39} with 30 volumes is the source domain data and the T2-SPIR MRI training 
data from ISBI 2019 CHAOS Challenge \cite{38} with 20 volumes is the target domain data. In each volume, a total 
of four abdominal organs are annotated, including liver, right (R.) kidney, left (L.) kidney, and spleen.

\noindent \textbf{Multi-modality Cardiac Segmentation.} We use Multi-Modality Whole Heart Segmentation (MMWHS) 2017 
Challenge dataset \cite{40}, which consists of 20 unaligned MRI and CT volumes with ground truth masks on each 
modality. We use the MRI and CT images as the source and target domain data, respectively. A total of four 
cardiac structures are annotated, including ascending aorta (AA), left atrium blood cavity (LAC), left 
ventricle blood cavity (LVC), and myocardium of the left ventricle (MYO). 

\begin{table*}[!t]
    \small
    \centering
    \caption{Quantitative results on cross-modality abdominal multi-organ segmentation. * represents our reproduced results.}
    \scalebox{0.95}{
      \begin{tabular}{c|ccccc|ccccc}
      \toprule
      \multirow{2}[4]{*}{Method} & \multicolumn{5}{c|}{Dice ↑}      & \multicolumn{5}{c}{ASD ↓} \\
  \cmidrule{2-11}          & Liver & R.kidney & L.kidney & Spleen & Average & Liver & R.kidney & L.kidney & Spleen & Average \\
      \midrule
      Supervised & 88.3  & 90.2  & 92.1  & 95.2  & 91.5  & 1.1   & 1.5   & 0.8   & 1.2   & 1.2  \\
      W/o adaptation & 55.1  & 41.7  & 54.3  & 62.2  & 53.3  & 4.3   & 9.8   & 5.3   & 7.9   & 6.8  \\
      \midrule
      SynSeg-Net (TMI 18') & 87.2  & \textbf{90.2}  & 76.6  & 79.6  & 83.4  & 2.8   & 0.7   & 4.8   & 2.5   & 2.7  \\
      AdaOutput (CVPR 18') & 85.8  & 89.7  & 76.3  & 82.2  & 83.5  & 1.9   & 1.4   & 3.0   & 1.8   & 2.1  \\
      CycleGAN (ICCV 17') & 88.8  & 87.3  & 76.8  & 79.4  & 83.1  & 2.0   & 3.2   & 1.9   & 2.6   & 2.4  \\
      CyCADA (ICML 18') & 88.7  & 89.3  & 78.1  & 80.2  & 84.1  & 1.5   & 1.7   & 1.3   &  \textbf{1.6}   & \textbf{1.5} \\
      Prior SIFA (AAAI 19') & 88.5  & 90.0 & 79.7  & 81.3  & 84.9  & 2.3   & 0.9   & 1.4   & 2.4   & 1.7  \\
      SIFA (TMI 20') & \textbf{90.0} & 89.1  & 80.2  & 82.3  & 85.4  & 1.5   & \textbf{0.6} & 1.5   & 2.4   & \textbf{1.5}  \\
      SIFA (TMI 20') * & 88.8  & 88.6  & 79.8  & 82.6  & 85.0  & 1.8   & 1.0   & 1.3   & 2.7   & 1.7  \\
      FDA (CVPR 20') * & 85.2  & 85.2  & 78.2  & 80.2  & 82.2  & 1.8   & 1.2   & 2.4   & 3.2   & 2.2  \\
      \midrule
      Ours  & 89.6  & 90.0 & \textbf{90.3} & \textbf{87.0} & \textbf{89.2} & \textbf{1.4} & 1.5   & \textbf{1.2} & 2.0 & \textbf{1.5} \\
      \bottomrule
      \end{tabular}%
    }
    \label{table1}%
  \end{table*}%

\begin{table*}[!t]
    \small
    \centering
    \caption{Quantitative results on cross-modality cardiac segmentation. * represents our reproduced results.}
    \scalebox{0.95}{
      \begin{tabular}{c|ccccc|ccccc}
      \toprule
      \multirow{2}[4]{*}{Method} & \multicolumn{5}{c|}{Dice  ↑}      & \multicolumn{5}{c}{ASD  ↓} \\
  \cmidrule{2-11}          & AA    & LAC   & LVC   & MYO   & Average & AA    & LAC   & LVC   & MYO   & Average \\
      \midrule
      Supervised & 94.1  & 91.9  & 94.5  & 85.7  & 91.6  & 1.5   & 0.7   & 1.4   & 1.5   & 1.3  \\
      W/o adaptation & 60.4  & 37.5  & 38.2  & 48.0  & 46.0  & 9.4   & 16.2  & 15.1  & 11.5  & 13.1  \\
      \midrule
      PnP-AdaNet (Access 19')  & 74.0  & 68.9  & 61.9  & 50.8  & 63.9  & 12.8  & 6.3   & 17.4  & 14.7  & 12.8  \\
      SynSeg-Net (TMI 18')  & 71.6  & 69.0  & 51.6  & 40.8  & 58.2  & 11.7  & 7.8   & 7.0   & 9.2   & 8.9  \\
      AdaOutput (CVPR 18')  & 65.2  & 76.6  & 54.4  & 43.6  & 59.9  & 17.9  & 5.5   & 5.9   & 8.9   & 9.6  \\
      CycleGAN (ICCV 17')  & 73.8  & 75.7  & 52.3  & 28.7  & 57.6  & 11.5  & 13.6  & 9.2   & 8.8   & 10.8  \\
      CyCADA (ICML 18')  & 72.9  & 77.0  & 62.4  & 45.3  & 64.4  & 9.6   & 8.0   & 9.6   & 10.5  & 9.4  \\
      Prior SIFA (AAAI 19')  & 81.1  & 76.4  & 75.7  & 58.7  & 73.0  & 10.6  & 7.4   & 6.7   & 7.8   & 8.1  \\
      SIFA (TMI 20')  & 81.3  & 79.5  & 73.8  & 61.6  & 74.1  & 7.9   & 6.2   & 5.5   & 8.5   & 7.0  \\
      FDA (CVPR 20') * & 80.8  & 76.2  & 75.9  & 60.5  & 73.4  & 8.2   & 5.8   & 9.2   & 9.3   & 8.1  \\
      ICMSC (MICCAI 21') & 85.6  & 86.4  & 84.3  & 72.4  & 82.2  & 2.4   & 3.3   & 3.4   & 3.2   & 3.1  \\
      CUDA (J-BHI 22') & \textbf{87.2} & \textbf{88.5}  & 83.0  & 72.8  & 82.9  & 7.0   & 2.8   & 5.2   & 6.8   & 5.5  \\
      \midrule
      Ours  & 86.8  & 87.5 & \textbf{84.6} & \textbf{82.4} & \textbf{85.3} & \textbf{1.6} & \textbf{2.5} & \textbf{3.2} & \textbf{3.1} & \textbf{2.6} \\
      \bottomrule
      \end{tabular}%
    }
    \label{table2}%
  \end{table*}%
  
  \noindent  \textbf{4.1.2 Evaluation Metrics.} We employed two commonly-used metrics to evaluate the segmentation performance: 
the Dice similarity coefficient (Dice) and the Average Symmetric Surface Distance (ASD).
  
\noindent  \textbf{4.1.3 Experimental Setting.}
We applied TransUnet \cite{37} as the backbone of the two teacher networks and the student network. Our network was 
trained on an NVIDIA GTX 3090 GPU with a batch size of 32 and 50 epoch, using an ADAM optimizer with a learning rate of 1e-4. We used the official data split for training and testing, 
and more details are described in  \textbf{Supplementary Material Section 1}.

\subsection{Comparison Results}
We compared our method with 10 existing state-of-the-art methods, including eight medical UDA methods: PnP-AdaNet \cite{16}, 
SynSeg-Net \cite{1}, CyCADA \cite{18}, Prior SIFA \cite{13}, SIFA \cite{2}, FDA \cite{19}, CUDA \cite{41}, ICMSC \cite{14} and two natural image 
UDA methods: AdaOutput \cite{20}, CycleGAN \cite{17}. Among them, we reproduced SIFA and FDA, and the results 
of the other studies are from their original papers. We also report the result of `W/o adaptation' as the lower 
bound of UDA segmentation performance, which is obtained by directly using the model trained on the source domain 
data to the target domain without adaptation. Similarly, we report the result of `Supervised' as the upper bound 
by training with labeled data in the target domain.

\begin{figure*}[htbp]
    \small
    \centering
    \includegraphics[scale=0.62]{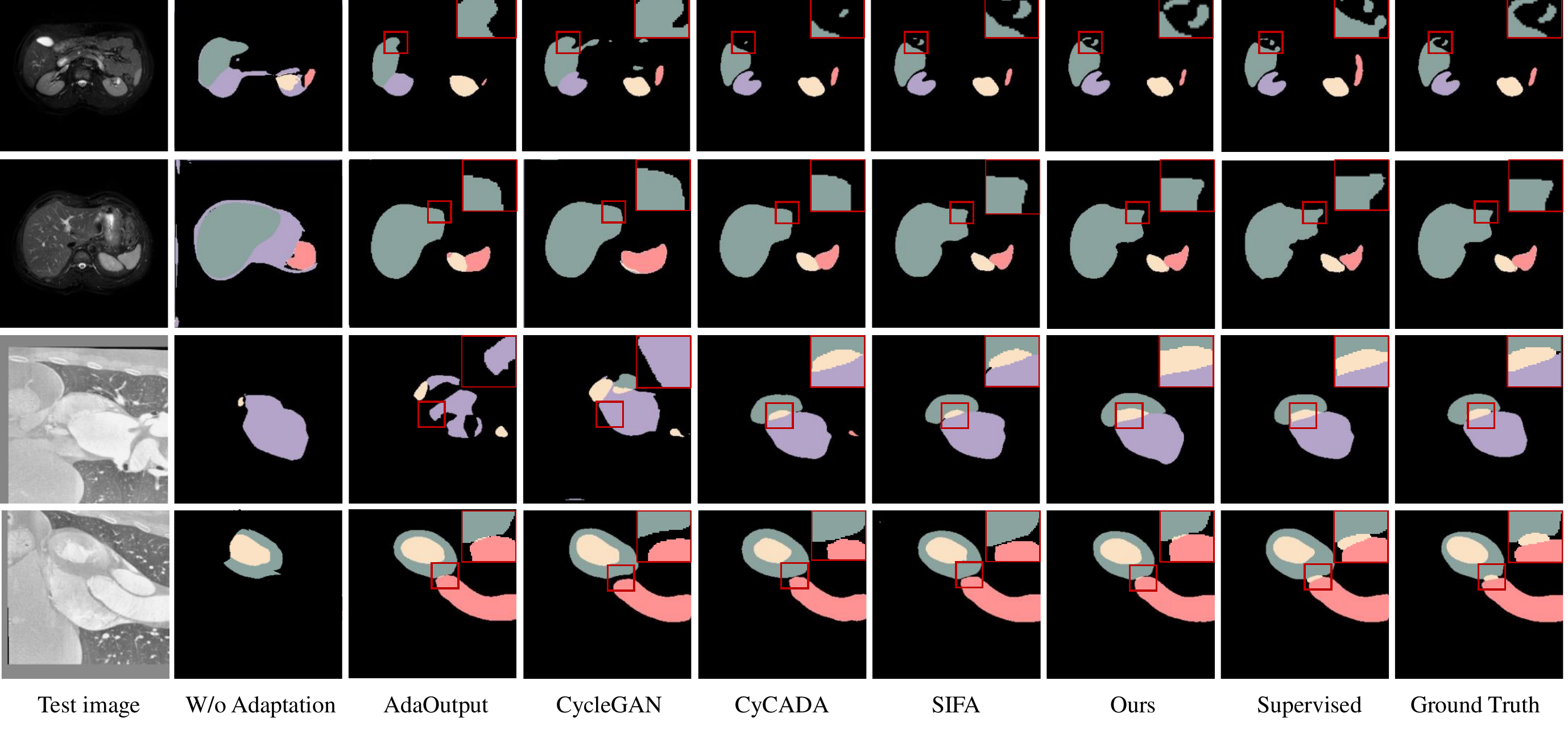}
    \caption{Visual comparison of segmentation results produced by different methods for abdominal 
    multi-organ MRI images (the first and second rows) and cardiac CT images (the third and fourth rows). 
    The liver, right kidney, left kidney and spleen are colored in blue, purple, yellow and red respectively. 
    The cardiac structures of AA, LAC, LVC and MYO are colored in red, purple, yellow and blue respectively.}
    \label{figure4}
\end{figure*}

\noindent  \textbf{4.2.1 Abdominal Multi-organ Image Segmentation}

\noindent The results of abdominal multi-organ segmentation are shown in Table \ref{table1}. Our proposed method achieves the best 
average Dice and average ASD, and it outperforms the comparing methods by a large margin in terms of average Dice. 
In the four organs, our method achieves the best performance on three of them in terms of both Dice 
and ASD. Without adaptation, the trained model can only achieve an average Dice of 53.3$\%$ and an average ASD 
of 6.8 on the target domain images. When our proposed UDA method is used, the Dice is increased significantly 
by 35.9$\%$ and the ASD is reduced by 5.3. Noteworthily, the average Dice of our method is only 2.3$\%$ 
lower than the supervised upper bound. The visualization results  are shown in Figure \ref{figure4} (Rows 1-2). 
Without adaptation, the network can barely predict correct abdominal organs. The segmentation 
results of AdaptOut and CycleGAN show much noise and incorrect prediction masks. Our proposed method produces more 
precise segmentation results than CyCADA and SIFA.

\noindent  \textbf{4.2.2 Multi-modality Cardiac Segmentation}

\noindent The results of multi-modality cardiac segmentation are shown in Table \ref{table2}. Again, our method achieves the 
best average Dice and average ASD and the margin over comparing method are fairly large in terms of both Dice 
and ASD. Compared to `W/o adaption', our method can increase the average Dice by 39.3$\%$ and reduce the average 
ASD by 10.5. Our method achieves the best performance on all four parts of heart in terms of ASD and on 
three parts in terms of Dice. The visualization results are shown in Figure \ref{figure4} (Rows 3-4). Compared 
with other UDA methods, our proposed method generates more accurate predicted cardiac structures.

\subsection{Ablation Studies on Key Components}


We conducted ablation studies on each key component proposed in our adaptation framework on both datasets and the results 
are shown in Table \ref{table3}, where `+ Spatial Transfer' and `+ Frequency Transfer' 
means training the segmentation model with spatial domain and frequency domain transferred images, respectively, 
and `+ S$\&$F Training' means mixing transferred images from both spatial and frequency domains to train the segmentation model. 
Compared with `W/o adaptation', both the spatial transfer and the frequency transfer 
strategy can significantly improve the segmentation performance on target domains. These results demonstrate the 
effectiveness of both transfer strategies. While simple mixing the spatial and frequency transferred images 
for model training is not effective, employing the multi-teacher knowledge distillation framework 
can improve the performance on target domains. These results indicate the effectiveness of the 
proposed multi-teacher distillation framework. The visualization of methods represent in \textbf{Supplementary Material Section 4}.

\begin{figure}[!t]
    \centering
    \includegraphics[scale=0.43]{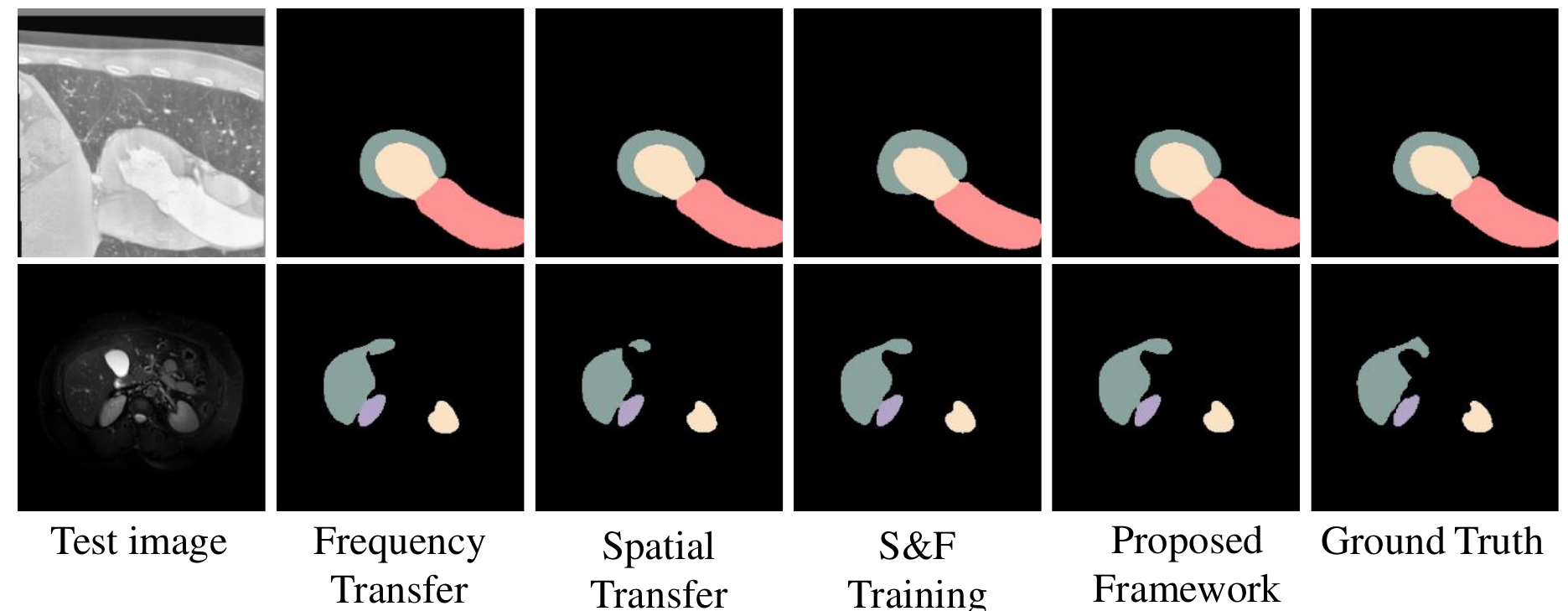}
    \caption{Visualization  results of cardiac parts and abdominal organs using 
    different settings in the ablation study.}
    \label{figure5}
\end{figure}

\begin{table}[!t]
    \small
    \centering
    \caption{The ablation studies on the effectiveness of key components. }
    \scalebox{0.9}{
      \begin{tabular}{c|cc|cc}
      \toprule
      \multirow{2}[4]{*}{Method} & \multicolumn{2}{c|}{Abdominal } & \multicolumn{2}{c}{Cardiac} \\
  \cmidrule{2-5}          & Dice ↑  & ASD ↓   & Dice ↑  & ASD ↓ \\
      \midrule
      W/o adaptation & 53.3  & 6.8   & 46.0  & 13.1  \\
      + Spatial Transfer(S) & 88.6  & \textbf{1.4}   & 83.7  & 3.2  \\
      + Frequency Transfer(F) & 87.3  & 2.0   & 84.6  & \textbf{2.4}  \\
      + S\&F Trainning & 88.4  & 1.9   & 83.9  & 3.6  \\
      Proposed Framework & \textbf{89.2}  & 1.5   & \textbf{85.3}  & 2.6  \\
      \bottomrule
      \end{tabular}%
    }
    \label{table3}%
  \end{table}%
Figure \ref{figure5} visualizes the segmentation results of one example in each dataset in the ablation study. 
Generally speaking, every UDA setting shows good segmentation results. With 
the proposed framework, the segmentation results are more similar to the ground truth.

\subsection{Ablation Study on Different Histogram Matching Strategies}
To reduce the style bias between the source and the target domains, we propose to use a histogram matching 
strategy based on batch momentum update to generate target-like images in the spatial domain. 
To validate the effectiveness of the batch momentum update-based histogram matching strategy, we conducted 
ablation experiments on different ways of target domain data selection in the histogram matching process. We 
experimented on four histogram calculation methods, including calculating the average 
histogram of the entire target domain data, randomly selecting the histogram of a single target domain image, 
calculating the average histogram of the current batch of target images and our proposed batch momentum update 
based histogram matching. The experiment results on the cardiac dataset shown in Table \ref{table4} 
indicate that our proposed method achieves the best segmentation performance on target domain.

\begin{table}[htbp]
    \small
    \centering
    \caption{Experimental results of different histogram matching strategies on cardiac datasets 
    (source domain: MRI, target domain: CT)}
    \scalebox{0.98}{
      \begin{tabular}{p{13em}cc}
      \toprule
      \multicolumn{1}{c}{Method} & Dice  ↑ & ASD ↓ \\
      \midrule
      \multicolumn{1}{c}{Entire target domain images} & 81.5  & 3.4 \\
      \multicolumn{1}{c}{Single target image} & 82.8  & \textbf{3.1} \\
      \multicolumn{1}{c}{Batch average-based} & 83.4  & 3.3 \\
      Batch momentum update-based & \textbf{83.7}  & 3.2 \\
      \bottomrule
      \end{tabular}%
    }
    \label{table4}%
  \end{table}%

\section{Conclusion}
In this paper, we propose a novel multi-teacher distillation framework for UDA based on frequency 
domain and spatial domain image transfer. In the frequency domain, we use NSCT to decompose images 
into 15 FCs and generate target-like images by combining selected FCs from the source and the target 
domain images for model training. In the spatial domain, we utilize histogram matching based on batch 
momentum update strategy to generate another set of target-like images for model training. Two teacher 
models are obtained with the frequency domain transferred images and the spatial domain transferred 
images, and a student model is then trained for inference under a multi-teacher distillation framework. 
Comparisons to existing methods and ablation studies validate the effectiveness of our proposed method. 
The idea of combining spatial and frequency domain transfer strategies in a multi-teacher distillation 
framework have the potential to be used in other UDA tasks.

\appendix

\section{Appendix}
\setcounter{table}{0}  
\setcounter{figure}{0}
\renewcommand{\thetable}{A\arabic{table}}
\renewcommand{\thefigure}{A\arabic{figure}}

\subsection{Implementation Details}

\subsubsection{A.1.1 Experimental Setting}

We applied TransUNet \cite{37} as the backbone of the two teacher networks and the student network. Our network was trained on an NVIDIA GTX 3090 
GPU with a batch size of 32 and 50 epoch, using an Adam optimizer and a cosine annealing learning rate adjustment strategy with a learning rate 
of 1e-4 and a weight decay of 5e-4. We used the official data split for training and testing on two commonly-used segmentation datasets. 
For these two datasets, we split both source and target domain datasets with 80\% data for training and 20\% for testing according to their 
original setting. When training the two teacher networks, the training set of the source domain with different transfer strategies was used to 
optimize the model and then test on target domain testing set. After finishing the training of the two teachers, we used the same dataset splitting 
for training on source domain training set and testing on target domain.

\subsubsection{A.1.2 Implementation Algorithms}
The specific implementation algorithms of two teacher models training phase and multi-teacher distillation process are shown in Algorithm 1 and Algorithm 2, 
respectively. 

\begin{algorithm}[!h]
	\renewcommand{\algorithmicrequire}{\textbf{Require:}}
	\renewcommand{\algorithmicensure}{\textbf{Output:}}
	\caption{Training process of the two teacher models.}
	\label{alg1}
	\begin{algorithmic}[1]
        \REQUIRE Source dataset $D_s=\{(x_i^s,y_i^s)\}_{i=1}^{N_s}$, target dataset $D_t=\{x_j^t\}_{j=1}^{N_t}$, model based on frequency 
        transfer $\xi _1$, model based on spatial transfer $\xi _2$. \\

        \% \textit{Frequency based Transfer Strategy}\\
        
        \FOR {epoch $<$ $N$}
            \FOR {$x_i^s \in D_s$}
                \STATE Randomly select a target domain image $x_j^t \in D_t$.\\
                \STATE $f_i^s=NSCT(x_i^s ),f_j^t=NSCT(x_j^t ).$
                \STATE Get frequency components $f_i^{s\to t}$ and image $x_{Fi}^{s\to t}$ after frequency transfer.
                \STATE Generate the predicted mask \\
                $\hat{y}_{Fi}^{s\to t}=\xi_1\left(x_{Fi}^{s\to t}\right)$.
                \STATE Optimize model $\xi_1$ with $\mathcal{L}_{Fre}$.
            \ENDFOR
        \ENDFOR \\

        \% \textit{Spatial based Transfer Strategy for Style alignment}\\
        \FOR {epoch $<$ $N$}
            \FOR {$i^{th}$ batch $\mathcal{X}_s=\{x_1^s, x_2^s, \ldots ,x_B^s\} \in D_S$ and $\mathcal{X}_t=\{x_1^t, x_2^t, \ldots ,x_B^t\} \in D_t$}
                \STATE Compute the target domain average grayscale image $\bar{x}_i^t  =1/B \sum_{j=1}^B x_j^t$ .\\
                \FOR{ $x_j^s \in \mathcal{X}_s$}
                    \STATE Generate transferred image with momentum histogram matching \\
                    $x_{Sj}^{s \to t}= match\_histogram(x_j^s,\bar{x}_i^t )$
                    \IF{$i > 1$}
                        \STATE Update $\bar{x}_i^t$ with momentum $\eta $:\\
                          $\bar{x}_i^t=\eta \bar{x}_i^t +(1-\eta) \bar{x}_{i-1}^t$
                    \ENDIF
                    \STATE Generate the predicted mask \\
                    $\hat{y}_{Sj}^{s\to t}=\xi_2 (x_{Sj}^{s \to t} )$;
                    \STATE Optimize model $\xi_2$ with $\mathcal{L}_{Spa}$.
                \ENDFOR
            \ENDFOR
        \ENDFOR
        \ENSURE Model $\xi_1$ based on frequency transfer and model $\xi_2$ based on histogram matching.
	\end{algorithmic}  
\end{algorithm}

\begin{algorithm}[!h]
	\renewcommand{\algorithmicrequire}{\textbf{Require:}}
	\renewcommand{\algorithmicensure}{\textbf{Output:}}
	\caption{Training process of multi-teacher distillation.}
	\label{alg2}
	\begin{algorithmic}[1]
        \REQUIRE Source dataset $D_s=\{(x_i^s,y_i^s)\}_{i=1}^{N_s}$, teacher model $\xi_1, \xi_2$, student model $\psi $. \\

        \FOR {epoch $<$ $N$}
            \FOR {$x_i^s \in D_s$}
                \STATE Generate the predicted masks \\
                $\hat{y}_{Fi}=\xi_1 (x_{Fi}^{s \to t} ),\hat{y}_{Si}=\xi_2 (x_{Si}^{s \to t} ), 
                \hat{y}_{i}=\psi (x_{i}^{s} )$\\ 
                \STATE Optimize the student model $\psi$ with $\mathcal{L}_{MKD}$. \\
            \ENDFOR
        \ENDFOR \\

        \ENSURE Student model $\psi$. 
	\end{algorithmic}  
\end{algorithm}

\subsection{Non-Subsampled Contourlet Transform}
Non-Subsampled Contourlet transform (NSCT) can efficiently capture multi-scale visual information of images by decomposing the image into high-pass sub-bands representing 
details and low-pass sub-bands representing structural information \cite{29, 33}. Concretely, NSCT consists of two parts: a 
Non-Subsampled Pyramid (NSP) and a Non-Subsampled Directional Filter Bank (NSDFB) shown in Figure \ref{figure1}, which are used to obtain 
multi-scale and multi-directional frequency components, respectively. We briefly introduce the three-layer NSCT used in this study here.

In NSP process, $H_0 (z)$ and $H_1 (z)$ represent low-pass filter and high-pass filter, respectively. For an input image $x$, NSP first 
decomposes it into a low-pass image $LS_1$ and a band-pass image $BS_1$ at the first layer using $H_0 (z)$ and $H_1 (z)$. $LS_1$ is further 
decomposed by NSP in layer two as $LS_2$ and $BS_2$. After $L$ times of NSP decompositions, the input image $x$ is decomposed into one 
low-pass image $LS_L$ and $L$ sub-band images $BS_l, l=1,2,…,L$.

\begin{figure*}[htpb]
    \centering
    \includegraphics[scale=0.40]{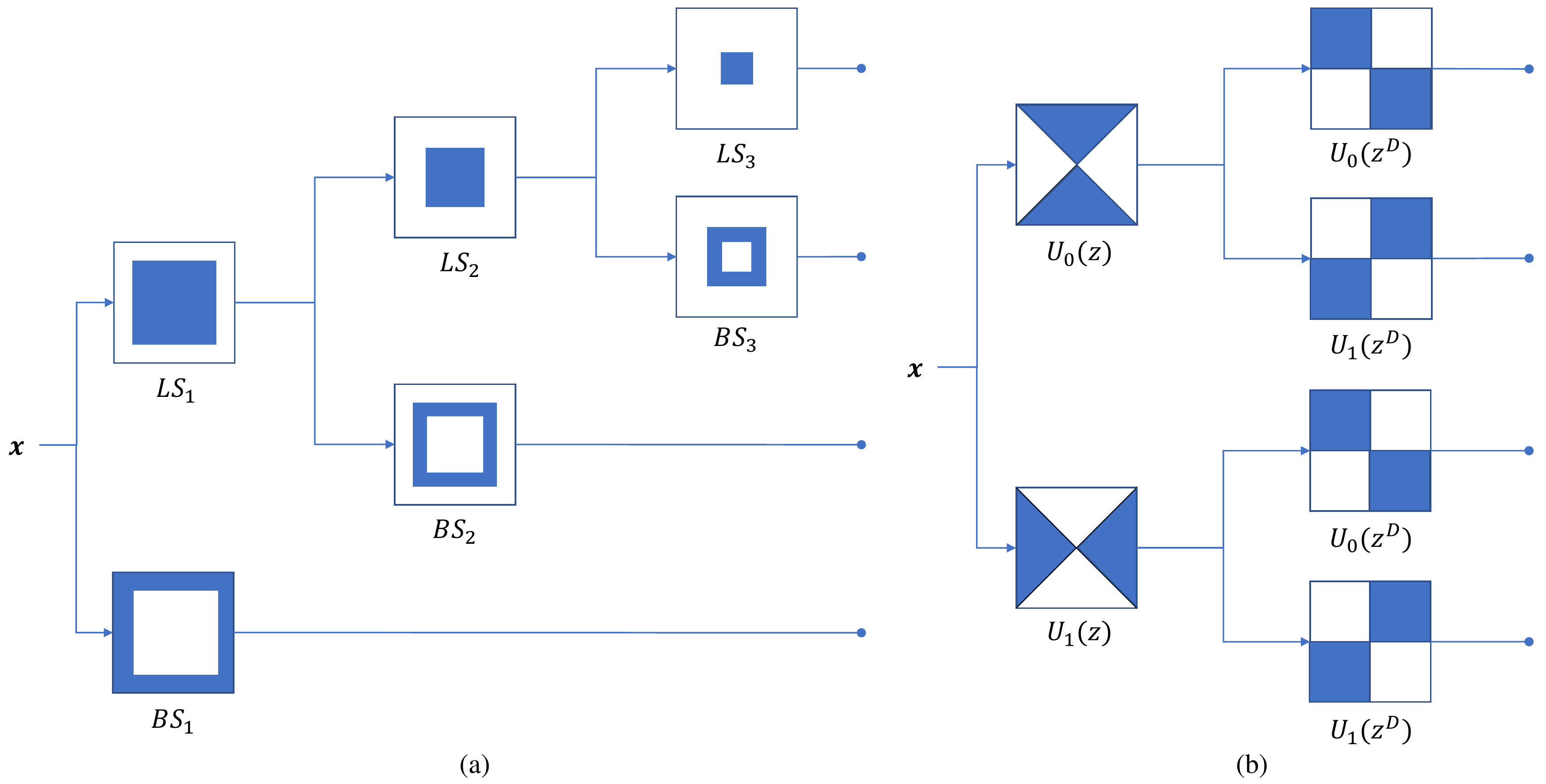}
    \caption{The details of NSP (left) and NSDFB (right).}
    \label{figure1}
\end{figure*}

NSDFB consists of non-subsampled directional filters including fan filters $U(z)$ and quadrature filters $U(z^D)$. For the band-pass image in $l^{th}$ 
layer, NSDFB decompose it into $2^l$ frequency components (FCs) in different directions. Thus, after $L$-level NSCT decomposition, the 
input image $x$ is transformed into a series of FCs, including $f(0)$ corresponding to the low-frequency sub-band $LS_L$, and band-pass 
FCs $f(1) \sim  f(\sum_{l=1}^L 2^l )$  corresponding to $L$ high frequency sub-bands $BS_l, l=1,2,…,L$.

\subsection{Empirical Experiments for DVFs and DIFs Identification}
In the pre-experiment, we perform NSCT and FCs transfer on the source and target domain images respectively, and then identify DIFs and DVFs according to the 
results using the transferred images. However, since the label of target domain images are not available, we match the histogram of the source domain image to 
that of the target ones to obtain target-like, labeled source domain image as "synthetic target domain data" for pre-experiment.

First, for each source domain image, we decompose it into 15 FCs using NSCT, and then reconstruct a series of images from different 
combinations of these FCs. The combination of frequency domain components is achieved by a band-stop filter, where the blocked frequency domain 
components are the `rejected FCs'. We train a baseline model 
using the original source domain images and train a transfer model using the images reconstructed from each FC combination. Then DVFs 
and DIFs are identified according to the performance of all these models on a synthetic target domain data. Specifically, an improved 
performance over the baseline model implies that the rejected FCs in the corresponding combination are domain-variant, and removing 
these FCs can prevent the model from learning too many domain-variant features to facilitate domain adaptation. On the contrary, a 
decreased performance over the baseline model implies that the removed FCs are domain-invariant, and by retaining these components, 
the model is encouraged to learn domain-invariant features to facilitate domain adaptation. In view of this insight, we conducted the 
experiments on synthetic target domain datasets because we have no access to segmentation ground truth in the target domain. Concretely, 
we use histogram matching strategy as described in original paper Section 3.3 to generate target style-like source images as synthetic 
target domain dataset. Then, we train the model using the new source images with different combination of FCs and test on synthetic target 
domain datasets in two applications.

The results of different FCs combinations on cardiac and abdominal multi-organ datasets are shown in Table \ref{tab1}.

\begin{table}[htbp]
    \small
    \centering
    \caption{Comparison results of different FCs combinations on cardiac and abdominal multi-organ datasets.}
    \scalebox{0.85}{
      \begin{tabular}{cccccc}
      \toprule
      \multicolumn{1}{c}{\multirow{2}[4]{*}{\textbf{No.}}} & \multirow{2}[4]{*}{\textbf{FCs}} & \multicolumn{2}{c}{\textbf{Cardiac Dice (\%) ↑}} & \multicolumn{2}{c}{\textbf{Abdominal Dice (\%) ↑}} \\
  \cmidrule{3-6}          & \multicolumn{1}{c}{} & \textbf{src val} & \multicolumn{1}{c}{\textbf{trgt val}} & \textbf{src val} & \multicolumn{1}{c}{\textbf{trgt val}} \\
      \midrule
      \multicolumn{1}{c}{Baseline} & \textcolor[rgb]{ .2,  .2,  .2}{[0,15]} & \textcolor[rgb]{ .2,  .2,  .2}{73.56} & \textcolor[rgb]{ .2,  .2,  .2}{60.42} & \textcolor[rgb]{ .2,  .2,  .2}{78.45} & \textcolor[rgb]{ .2,  .2,  .2}{68.25} \\
      1     & [0,1] & 49.95 & 41.15 & 58.65 & 44.26 \\
      2     & [1,3] & 67.74 & 58.23 & 65.25 & 52.32 \\
      3     & [3,7] & 75.13 & 58.56 & 79.61 & 67.25 \\
      4     & [7,14] & 71.54 & 62.65 & 75.74 & 71.63 \\
      5     & [0,1], [1,3] & 72.02 & 59.98 & 72.84 & 64.62 \\
      6     & [0,1], [3,7] & 70.16 & 60.25 & 73.69 & 68.86 \\
      7     & [0,1], [7,14] & \textcolor[rgb]{ .2,  .2,  .2}{\textbf{77.32}} & \textcolor[rgb]{ .2,  .2,  .2}{\textbf{64.85}} & \textcolor[rgb]{ .2,  .2,  .2}{\textbf{80.17}} & \textcolor[rgb]{ .2,  .2,  .2}{\textbf{72.54}} \\
      \bottomrule
      \end{tabular}%
      }
    \label{tab1}%
  \end{table}%

From Table \ref{tab1}, we can see that compared to Baseline (full FCs), when we keep the low-frequency FC ([0,1]) and the third-layer FCs ([7,14]), 
the segmentation results on synthetic target domain validation dataset achieve improved and best performance on both applications. Therefore, 
we consider these FCs as DIFs and the rest as DVFs. The low-frequency component represents the shape and semantics of the image, which is 
important for semantic information learning. However, the segmentation results of only preserving low-frequency component show poor performance, 
which demonstrates the significance of the high-frequency components containing details and textures. When we keep different high-level FCs 
combined with low-level FC, the performance is consistently improved compared with the results only with different high-level FCs and only 
with low-level FC, which both validates the importance of semantics containing in low-level FC and details containing in high-level FCs.

\subsection{Visualization results of Different Transfer Strategies}
Examples of adapted images with different transfer strategies are visualized in Figure \ref{figure2}. We can see that the appearance of the source 
images is indeed adapted into target domains in different perspectives. In the frequency domain, we are expected to randomize the domain-variant 
frequency components while keeping the domain-invariant components unchanged. In the spatial domain, we are expected to align their style bias 
between different domains.

\begin{figure}[t]
    
    \centering
    \includegraphics[scale=0.60]{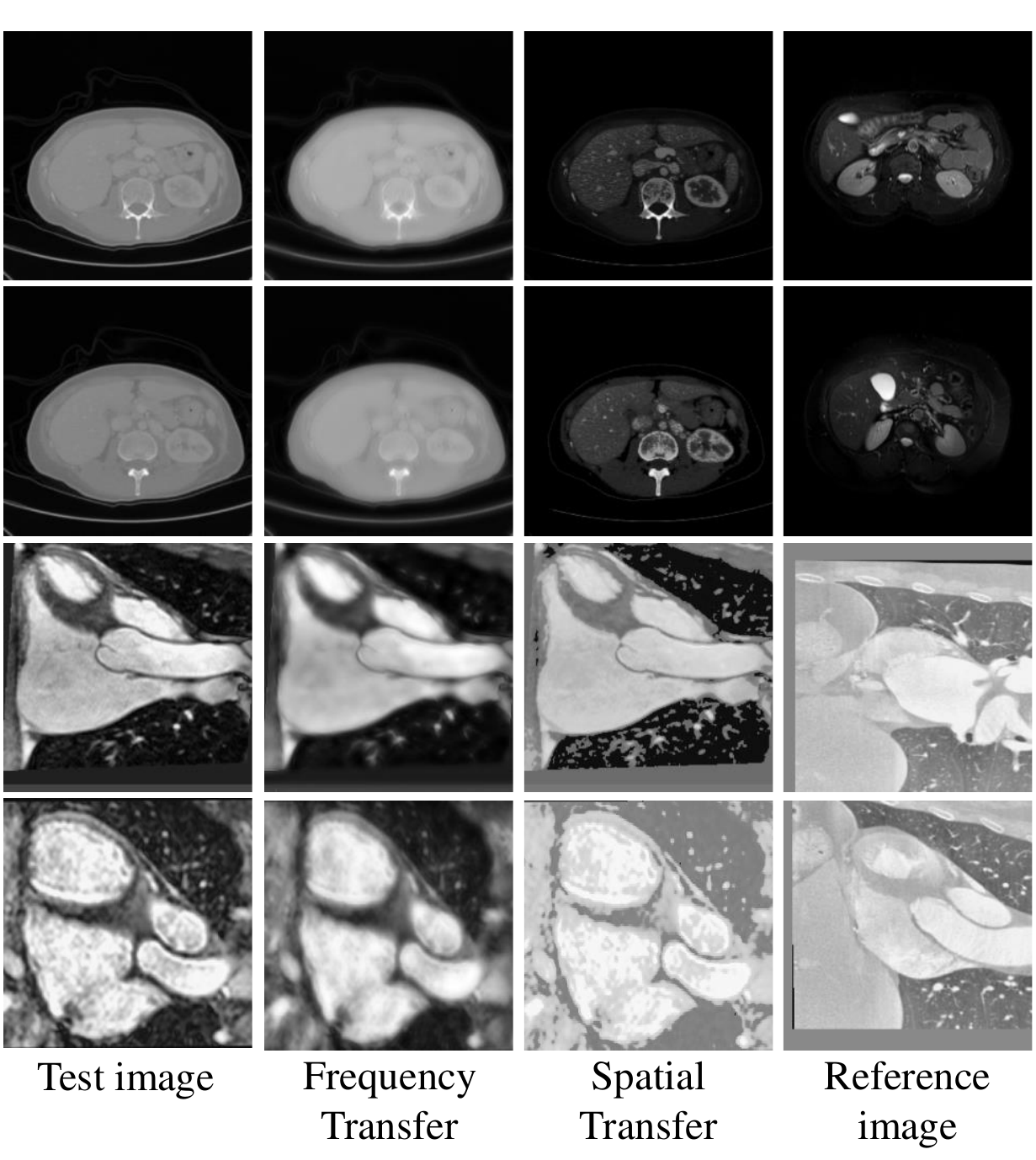}
    \caption{Examples of adapted images with different transfer strategies. The first two rows are from chaos CT images, and the last two rows 
    are from cardiac MRI images.}
    \label{figure2}
\end{figure}

\bibliography{aaai23}

\begin{thebibliography}{35}
\providecommand{\natexlab}[1]{#1}

\bibitem[{AlBadawy, Saha, and Mazurowski(2018)}]{10}
AlBadawy, E.~A.; Saha, A.; and Mazurowski, M.~A. 2018.
\newblock Deep learning for segmentation of brain tumors: Impact of
  cross-institutional training and testing.
\newblock \emph{Medical Physics}, 45(3): 1150--1158.

\bibitem[{Bhatnagar, Wu, and Liu(2013)}]{33}
Bhatnagar, G.; Wu, Q.~J.; and Liu, Z. 2013.
\newblock Directive contrast based multimodal medical image fusion in NSCT
  domain.
\newblock \emph{IEEE Transactions on Multimedia}, 15(5): 1014--1024.

\bibitem[{Chen et~al.(2019)Chen, Dou, Chen, Qin, and Heng}]{13}
Chen, C.; Dou, Q.; Chen, H.; Qin, J.; and Heng, P.-A. 2019.
\newblock Synergistic image and feature adaptation: Towards cross-modality
  domain adaptation for medical image segmentation.
\newblock In \emph{Proceedings of the AAAI Conference on Artificial
  Intelligence}, volume~33, 865--872.

\bibitem[{Chen et~al.(2020)Chen, Dou, Chen, Qin, and Heng}]{2}
Chen, C.; Dou, Q.; Chen, H.; Qin, J.; and Heng, P.~A. 2020.
\newblock Unsupervised bidirectional cross-modality adaptation via deeply
  synergistic image and feature alignment for medical image segmentation.
\newblock \emph{IEEE Transactions on Medical Imaging}, 39(7): 2494--2505.

\bibitem[{Chen et~al.(2021{\natexlab{a}})Chen, Lu, Yu, Luo, Adeli, Wang, Lu,
  Yuille, and Zhou}]{37}
Chen, J.; Lu, Y.; Yu, Q.; Luo, X.; Adeli, E.; Wang, Y.; Lu, L.; Yuille, A.~L.;
  and Zhou, Y. 2021{\natexlab{a}}.
\newblock Transunet: Transformers make strong encoders for medical image
  segmentation.
\newblock \emph{arXiv preprint arXiv:2102.04306}.

\bibitem[{Chen et~al.(2021{\natexlab{b}})Chen, Jia, He, Shi, and Liu}]{22}
Chen, S.; Jia, X.; He, J.; Shi, Y.; and Liu, J. 2021{\natexlab{b}}.
\newblock Semi-supervised domain adaptation based on dual-level domain mixing
  for semantic segmentation.
\newblock In \emph{Proceedings of the IEEE/CVF Conference on Computer Vision
  and Pattern Recognition}, 11018--11027.

\bibitem[{Csurka et~al.(2017)}]{8}
Csurka, G.; et~al. 2017.
\newblock \emph{Domain adaptation in computer vision applications}.
\newblock Springer.

\bibitem[{Dou et~al.(2018)Dou, Ouyang, Chen, Chen, Glocker, Zhuang, and
  Heng}]{16}
Dou, Q.; Ouyang, C.; Chen, C.; Chen, H.; Glocker, B.; Zhuang, X.; and Heng,
  P.-A. 2018.
\newblock Pnp-adanet: Plug-and-play adversarial domain adaptation network with
  a benchmark at cross-modality cardiac segmentation.
\newblock \emph{arXiv preprint arXiv:1812.07907}.

\bibitem[{Du and Liu(2021)}]{41}
Du, X.; and Liu, Y. 2021.
\newblock Constraint-Based Unsupervised Domain Adaptation Network for
  Multi-Modality Cardiac Image Segmentation.
\newblock \emph{IEEE Journal of Biomedical and Health Informatics}, 26(1):
  67--78.

\bibitem[{Ghafoorian et~al.(2017)Ghafoorian, Mehrtash, Kapur, Karssemeijer,
  Marchiori, Pesteie, Guttmann, Leeuw, Tempany, Ginneken et~al.}]{11}
Ghafoorian, M.; Mehrtash, A.; Kapur, T.; Karssemeijer, N.; Marchiori, E.;
  Pesteie, M.; Guttmann, C.~R.; Leeuw, F.-E.~d.; Tempany, C.~M.; Ginneken,
  B.~v.; et~al. 2017.
\newblock Transfer learning for domain adaptation in MRI: Application in brain
  lesion segmentation.
\newblock In \emph{International Conference on Medical Image Computing and
  Computer-assisted Intervention}, 516--524. Springer.

\bibitem[{Gonzalez(2009)}]{35}
Gonzalez, R.~C. 2009.
\newblock \emph{Digital image processing}.
\newblock Pearson Education India.

\bibitem[{Guan and Liu(2021)}]{9}
Guan, H.; and Liu, M. 2021.
\newblock Domain adaptation for medical image analysis: a survey.
\newblock \emph{IEEE Transactions on Biomedical Engineering}, 69(3):
  1173--1185.

\bibitem[{He et~al.(2021)He, Jia, Chen, and Liu}]{23}
He, J.; Jia, X.; Chen, S.; and Liu, J. 2021.
\newblock Multi-source domain adaptation with collaborative learning for
  semantic segmentation.
\newblock In \emph{Proceedings of the IEEE/CVF Conference on Computer Vision
  and Pattern Recognition}, 11008--11017.

\bibitem[{Hoffman et~al.(2018)Hoffman, Tzeng, Park, Zhu, Isola, Saenko, Efros,
  and Darrell}]{18}
Hoffman, J.; Tzeng, E.; Park, T.; Zhu, J.-Y.; Isola, P.; Saenko, K.; Efros, A.;
  and Darrell, T. 2018.
\newblock Cycada: Cycle-consistent adversarial domain adaptation.
\newblock In \emph{International Conference on Machine Learning}, 1989--1998.
  Pmlr.

\bibitem[{Huang et~al.(2021)Huang, Guan, Xiao, and Lu}]{27}
Huang, J.; Guan, D.; Xiao, A.; and Lu, S. 2021.
\newblock Fsdr: Frequency space domain randomization for domain generalization.
\newblock In \emph{Proceedings of the IEEE/CVF Conference on Computer Vision
  and Pattern Recognition}, 6891--6902.

\bibitem[{Huo et~al.(2018)Huo, Xu, Moon, Bao, Assad, Moyo, Savona, Abramson,
  and Landman}]{1}
Huo, Y.; Xu, Z.; Moon, H.; Bao, S.; Assad, A.; Moyo, T.~K.; Savona, M.~R.;
  Abramson, R.~G.; and Landman, B.~A. 2018.
\newblock Synseg-net: Synthetic segmentation without target modality ground
  truth.
\newblock \emph{IEEE Transactions on Medical Imaging}, 38(4): 1016--1025.

\bibitem[{Kavur et~al.(2021)Kavur, Gezer, Bar{\i}{\c{s}}, Aslan, Conze, Groza,
  Pham, Chatterjee, Ernst, {\"O}zkan et~al.}]{38}
Kavur, A.~E.; Gezer, N.~S.; Bar{\i}{\c{s}}, M.; Aslan, S.; Conze, P.-H.; Groza,
  V.; Pham, D.~D.; Chatterjee, S.; Ernst, P.; {\"O}zkan, S.; et~al. 2021.
\newblock CHAOS challenge-combined (CT-MR) healthy abdominal organ
  segmentation.
\newblock \emph{Medical Image Analysis}, 69: 101950.

\bibitem[{Kwon et~al.(2020)Kwon, Na, Lee, and Kim}]{34}
Kwon, K.; Na, H.; Lee, H.; and Kim, N.~S. 2020.
\newblock Adaptive knowledge distillation based on entropy.
\newblock In \emph{ICASSP 2020-2020 IEEE International Conference on Acoustics,
  Speech and Signal Processing}, 7409--7413. IEEE.

\bibitem[{Landman et~al.(2015)Landman, Xu, Igelsias, Styner, Langerak, and
  Klein}]{39}
Landman, B.; Xu, Z.; Igelsias, J.; Styner, M.; Langerak, T.; and Klein, A.
  2015.
\newblock Miccai multi-atlas labeling beyond the cranial vault--workshop and
  challenge.
\newblock In \emph{Proc. MICCAI Multi-Atlas Labeling Beyond Cranial
  Vault—Workshop Challenge}, volume~5, 12.

\bibitem[{Ma et~al.(2021)Ma, Lin, Wu, and Yu}]{25}
Ma, H.; Lin, X.; Wu, Z.; and Yu, Y. 2021.
\newblock Coarse-to-fine domain adaptive semantic segmentation with photometric
  alignment and category-center regularization.
\newblock In \emph{Proceedings of the IEEE/CVF Conference on Computer Vision
  and Pattern Recognition}, 4051--4060.

\bibitem[{Ma et~al.(2022)Ma, Yuan, Chen, Tong, and Lin}]{21}
Ma, X.; Yuan, J.; Chen, Y.-w.; Tong, R.; and Lin, L. 2022.
\newblock Attention-based cross-layer domain alignment for unsupervised domain
  adaptation.
\newblock \emph{Neurocomputing}, 499: 1--10.

\bibitem[{Nam et~al.(2021)Nam, Lee, Park, Yoon, and Yoo}]{24}
Nam, H.; Lee, H.; Park, J.; Yoon, W.; and Yoo, D. 2021.
\newblock Reducing domain gap by reducing style bias.
\newblock In \emph{Proceedings of the IEEE/CVF Conference on Computer Vision
  and Pattern Recognition}, 8690--8699.

\bibitem[{Shen et~al.(2020)Shen, Yao, Yan, Tian, Jiang, and Zhou}]{6}
Shen, R.; Yao, J.; Yan, K.; Tian, K.; Jiang, C.; and Zhou, K. 2020.
\newblock Unsupervised domain adaptation with adversarial learning for mass
  detection in mammogram.
\newblock \emph{Neurocomputing}, 393: 27--37.

\bibitem[{Toldo et~al.(2020)Toldo, Maracani, Michieli, and Zanuttigh}]{12}
Toldo, M.; Maracani, A.; Michieli, U.; and Zanuttigh, P. 2020.
\newblock Unsupervised domain adaptation in semantic segmentation: a review.
\newblock \emph{Technologies}, 8(2): 35.

\bibitem[{Tsai et~al.(2018)Tsai, Hung, Schulter, Sohn, Yang, and
  Chandraker}]{20}
Tsai, Y.-H.; Hung, W.-C.; Schulter, S.; Sohn, K.; Yang, M.-H.; and Chandraker,
  M. 2018.
\newblock Learning to adapt structured output space for semantic segmentation.
\newblock In \emph{Proceedings of the IEEE/CVF Conference on Computer Vision
  and Pattern Recognition}, 7472--7481.

\bibitem[{Wollmann, Eijkman, and Rohr(2018)}]{3}
Wollmann, T.; Eijkman, C.; and Rohr, K. 2018.
\newblock Adversarial domain adaptation to improve automatic breast cancer
  grading in lymph nodes.
\newblock In \emph{2018 IEEE 15th International Symposium on Biomedical
  Imaging}, 582--585. IEEE.

\bibitem[{Xing et~al.(2020)Xing, Cornish, Bennett, and Ghosh}]{5}
Xing, F.; Cornish, T.~C.; Bennett, T.~D.; and Ghosh, D. 2020.
\newblock Bidirectional mapping-based domain adaptation for nucleus detection
  in cross-modality microscopy images.
\newblock \emph{IEEE Transactions on Medical Imaging}, 40(10): 2880--2896.

\bibitem[{Xu et~al.(2021)Xu, Zhang, Zhang, Wang, and Tian}]{26}
Xu, Q.; Zhang, R.; Zhang, Y.; Wang, Y.; and Tian, Q. 2021.
\newblock A fourier-based framework for domain generalization.
\newblock In \emph{Proceedings of the IEEE/CVF Conference on Computer Vision
  and Pattern Recognition}, 14383--14392.

\bibitem[{Yang and Soatto(2020)}]{19}
Yang, Y.; and Soatto, S. 2020.
\newblock Fda: Fourier domain adaptation for semantic segmentation.
\newblock In \emph{Proceedings of the IEEE Conference on Computer Vision and
  Pattern Recognition}, 4085--4095.

\bibitem[{Yaras et~al.(2021)Yaras, Huang, Bradbury, and Malof}]{36}
Yaras, C.; Huang, B.; Bradbury, K.; and Malof, J.~M. 2021.
\newblock Randomized Histogram Matching: A Simple Augmentation for Unsupervised
  Domain Adaptation in Overhead Imagery.
\newblock \emph{arXiv preprint arXiv:2104.14032}.

\bibitem[{Zeng et~al.(2021)Zeng, Lerch, Schmaranzer, Zheng, Burger, Gerber,
  Tannast, Siebenrock, and Gerber}]{14}
Zeng, G.; Lerch, T.~D.; Schmaranzer, F.; Zheng, G.; Burger, J.; Gerber, K.;
  Tannast, M.; Siebenrock, K.; and Gerber, N. 2021.
\newblock Semantic consistent unsupervised domain adaptation for cross-modality
  medical image segmentation.
\newblock In \emph{International Conference on Medical Image Computing and
  Computer Assisted Intervention}, 201--210. Springer.

\bibitem[{Zhang et~al.(2019)Zhang, Liu, Pan, and Shen}]{4}
Zhang, J.; Liu, M.; Pan, Y.; and Shen, D. 2019.
\newblock Unsupervised conditional consensus adversarial network for brain
  disease identification with structural MRI.
\newblock In \emph{International Workshop on Machine Learning in Medical
  Imaging}, 391--399. Springer.

\bibitem[{Zhu et~al.(2017)Zhu, Park, Isola, and Efros}]{17}
Zhu, J.-Y.; Park, T.; Isola, P.; and Efros, A.~A. 2017.
\newblock Unpaired image-to-image translation using cycle-consistent
  adversarial networks.
\newblock In \emph{Proceedings of the IEEE International Conference on Computer
  Vision}, 2223--2232.

\bibitem[{Zhu et~al.(2019)Zhu, Zheng, Qi, Wang, and Xiang}]{29}
Zhu, Z.; Zheng, M.; Qi, G.; Wang, D.; and Xiang, Y. 2019.
\newblock A Phase Congruency and Local Laplacian Energy Based Multi-Modality
  Medical Image Fusion Method in NSCT Domain.
\newblock \emph{IEEE Access}, 7: 20811--20824.

\bibitem[{Zhuang and Shen(2016)}]{40}
Zhuang, X.; and Shen, J. 2016.
\newblock Multi-scale patch and multi-modality atlases for whole heart
  segmentation of MRI.
\newblock \emph{Medical Image Analysis}, 31: 77--87.

\end{thebibliography}

\end{document}